\documentclass[journal]{IEEEtranTIE}
\usepackage{graphicx}
\usepackage{cite}
\usepackage{picinpar}
\usepackage{amsmath, amsfonts}
\usepackage{url}
\usepackage{flushend}
\usepackage[latin1]{inputenc}
\usepackage{colortbl}
\usepackage{soul}
\usepackage{multirow}
\usepackage{booktabs}
\usepackage{threeparttable}
\usepackage{dcolumn}
\usepackage{bm}
\usepackage{pifont}
\usepackage{color}
\usepackage{alltt}
\usepackage{array}
\usepackage{textcomp}
\usepackage{stfloats}
\usepackage{verbatim}
\usepackage{enumerate}
\usepackage{siunitx}
\usepackage{breakurl}
\usepackage{epstopdf}
\usepackage{pbox}
\usepackage{subcaption}
\usepackage{caption}
\usepackage{float}
\usepackage[colorlinks,urlcolor=blue,linkcolor=blue,citecolor=blue]{hyperref}

\begin{document}
\title{Grounded Vision-Language Navigation for UAVs with Open-Vocabulary Goal Understanding}

\author{
	\vskip 1em
	
	Yuhang Zhang, \emph{Student Member, IEEE},
        Haosheng Yu,  \emph{Student Member, IEEE},
        Jiaping Xiao, \emph{Student Member, IEEE},
	\\ and Mir Feroskhan, \emph{Member, IEEE}

	\thanks{
	
		% Manuscript received Month xx, 2xxx; revised Month xx, xxxx; accepted Month x, xxxx.
		% This work was supported in part by the xxx Department of xxx under Grant  (sponsor and financial support acknowledgment goes here).
		
        Yuhang Zhang, Haosheng Yu, Jiaping Xiao, and Mir Feroskhan are with the School of Mechanical and Aerospace Engineering, Nanyang Technological University, Singapore 639798, Singapore (e-mail: yuhang004@e.ntu.edu.sg; yuha0014@e.ntu.edu.sg; jiaping001@e.ntu.edu.sg; mir.feroskhan@ntu.edu.sg).
        Yuhang Zhang and Haosheng Yu are co-first authors. \textit{(Corresponding authors: Jiaping Xiao and Mir Feroskhan.)}

        Project materials and supplementary information are available at: \url{https://zzzzzyh111.github.io/VLFly/}.

	}
}

\maketitle
	
\begin{abstract}
Vision-and-language navigation (VLN) is a long-standing challenge in autonomous robotics, aiming to empower agents with the ability to follow human instructions while navigating complex environments. Two key bottlenecks remain in this field: generalization to out-of-distribution environments and reliance on fixed discrete action spaces. To address these challenges, we propose Vision-Language Fly (VLFly), a framework tailored for Unmanned Aerial Vehicles (UAVs) to execute language-guided flight. Without the requirement for localization or active ranging sensors, VLFly outputs continuous velocity commands purely from egocentric observations captured by an onboard monocular camera. The VLFly integrates three modules: an instruction encoder based on a large language model (LLM) that reformulates high-level language into structured prompts, a goal retriever powered by a vision-language model (VLM) that matches these prompts to goal images via vision-language similarity, and a waypoint planner that generates executable trajectories for real-time UAV control. VLFly is evaluated across diverse simulation environments without additional fine-tuning and consistently outperforms all baselines. Moreover, real-world VLN tasks in indoor and outdoor environments under direct and indirect instructions demonstrate that VLFly achieves robust open-vocabulary goal understanding and generalized navigation capabilities, even in the presence of abstract language input.

% Given the high maneuverability and limited stability of onboard visual sensing in aerial platforms, we argue that VLFly marks a meaningful step forward toward grounded, instruction-driven aerial navigation.
\end{abstract}


% \markboth{IEEE TRANSACTIONS ON INDUSTRIAL ELECTRONICS}%
% {}

\definecolor{limegreen}{rgb}{0.2, 0.8, 0.2}
\definecolor{forestgreen}{rgb}{0.13, 0.55, 0.13}
\definecolor{greenhtml}{rgb}{0.0, 0.5, 0.0}

\section{Introduction}
The capability of robots to execute complex tasks based on natural language instructions has long been a compelling objective in robotics and artificial intelligence (AI). Particularly within the field of autonomous navigation, this capability enables applications spanning home assistance \cite{zitkovich2023rt}, urban inspection \cite{feng2025citybench}, and environmental exploration \cite{devarakonda2024orionnav}. In such scenarios, robots are required to identify target objects belonging to a specified category and navigate efficiently toward them by understanding language instructions and perceiving visual observations, a challenging task known as vision-language navigation (VLN). For instance, suppose a pink plush toy is missing in the room, and the instruction "\textit{find a pink toy}" is given, the robot is not supposed to conduct an exhaustive exploration of the entire environment. Instead, it should intelligently direct its attention toward places of interest presenting visual cues strongly correlated with the language instruction. 
To effectively address such tasks, a competent VLN framework should demonstrate the following capabilities: (1) pixel-level interpretation of vision-language features; (2) accurate mapping of fused vision-language features into executable actions that are grounded within the environment; and (3) robust generalization to unstructured and previously unseen environments.

However, no existing VLN system fully realizes the aforementioned capabilities, particularly in the context of unmanned aerial vehicles (UAVs) operating in real-world environments. Existing autonomous navigation research can be broadly categorized into three main streams: traditional, end-to-end learning, and modular learning approaches \cite{gervet2023navigating}. Traditional methods \cite{mur2015orb, matsuki2024gaussian, cui2023mcsfm, wang2024tc}, such as Simultaneous Localization and Mapping (SLAM) and Structure from Motion (SfM) based navigation, focus on pixel-level feature extraction and matching for geometric reconstruction. Although effective for autonomous navigation, they offer no solution for visual-language feature integration and semantic understanding. End-to-end learning methods  \cite{ross2013learning, yan2021deep, loquercio2021learning, yan2023collision, xiao2023collaborative, kaufmann2023champion, xiao2024learning}, which rely on backpropagation of losses from Imitation Learning (IL) or Reinforcement Learning (RL) to enable pixel-to-action training, demonstrate promising performance but face challenges such as over-reliance on training data, inefficient sampling, overfitting in simulation, and large sim-to-real gaps. Modular learning approaches \cite{wu2022image, gervet2023navigating, roth2024viplanner} retain the structured pipeline of traditional methods while replacing specific computational modules with learned ones, thereby exhibiting improved generalization and robust real-world performance. However, they still require large-scale real-world datasets and suffer from module error accumulation and limited human-like reasoning. 

Recently, the emergence of Large Language Models (LLMs) \cite{touvron2023llama} and Vision-Language Models (VLMs) \cite{radford2021learning} has further advanced the development of autonomous navigation, and these models can also be incorporated into either end-to-end or modular learning frameworks. For end-to-end learning \cite{zhang2024navid, xu2025flame, zhang2025mapnav}, VLMs can extract semantically grounded features directly from RGB images and leverage language-annotated ground truth trajectories as direct supervisory signals to map these features to actions. This not only enhances sampling efficiency compared to conventional end-to-end learning methods but also enables open-vocabulary goal understanding. For modular learning \cite{yokoyama2024vlfm, wu2024voronav, du2025vl}, the visual-language features extracted by VLMs can be used to construct semantic maps, thereby facilitating exploration guided by human-level semantic cues. 
However, recent VLN frameworks  predominantly focus on ground-based agents, while UAV-based VLN research remains relatively underexplored due to the challenges of more complex action and observation space, even though UAVs offer a wider range of operating scenarios. Specifically, existing ground-based VLN methods typically assume a predefined discrete action space, which differs significantly from the continuous and dynamic nature of UAV flight. Moreover, occluded views and unstable visual observation in flight bring the environmental and perceptual gaps between ground and aerial navigation settings, hence limiting the direct transferability of ground-based VLN methods to UAVs.

% We argue that the challenges of UAV-based VLN are twofold: 
% \begin{itemize}
%     \item \textbf{Discrepancy in agent motion dynamics}. Ground-based agents typically operate on a two-dimensional plane, where it is intuitive and effective to control translation and rotation using discrete actions (e.g., moving forward \SI{15}{\centi\meter} or rotating \ang{15} clockwise). In contrast, UAVs perform complex, continuous maneuvers in three-dimensional space, such as diving, ascending diagonally, or executing roll-based turns. Representing UAV spatial movements using a discrete action space oversimplifies control and fails to reflect real-world flight behavior.
%     \item \textbf{Variation in agent perception}. Unlike ground-based agents, which usually operate in constrained indoor settings, UAVs navigate expansive outdoor environments and must cope with occluded views and unstable visual inputs resulting from their high agility and frequent viewpoint shifts. As a result, existing ground-based VLN paradigms are not directly transferable to UAVs.
% \end{itemize}

To address these challenges, we propose Vision-Language Fly (VLFly), a novel VLN framework designed for UAVs, capable of open-vocabulary goal understanding and zero-shot transfer in indoor and outdoor scenarios. VLFly integrates three modules to enable generalizable UAV navigation grounded in natural language instructions. Firstly, VLFly leverages an LLM to interpret high-level language instructions and transform them into structured prompts that enhance semantic alignment with visual representations. It then employs a VLM to perform goal grounding by retrieving a semantically aligned image from a pre-collected pool through cross-modal similarity matching. Lastly, once the goal is identified, VLFly predicts a sequence of navigational waypoints  based solely on egocentric visual observations along with the grounded goal image. These waypoints are subsequently converted into continuous control commands (e.g., linear and angular velocities) via a classical planner for real-time UAV execution. 
Rather than relying on handcrafted rules or task-specific fine-tuning, VLFly integrates pixel-level vision-language features within a unified, UAV-oriented architecture, leveraging temporally stacked egocentric observations and goal images to predict future waypoint trajectories for spatial planning. The framework outperforms all baselines, achieving open-vocabulary goal understanding and robust navigation performance across various simulated and real-world environments.
Our main contributions are summarized as follows:
\begin{enumerate}
    \item We propose VLFly, a novel and unified VLN framework specifically designed for UAVs, enabling open-vocabulary goal understanding and zero-shot transfer without the need for task-specific fine-tuning.
    \item We design modular VLN components for natural language understanding, cross-modal grounding, and navigable waypoint generation, effectively bridging the gap between semantic instructions and continuous UAV control commands.
    \item We conduct extensive evaluations in diverse simulated and real-world settings, demonstrating that VLFly achieves robust generalization and outperforms all baselines in UAV-based VLN tasks.
\end{enumerate}

The rest of this paper is organized as follows. Section \uppercase\expandafter{\romannumeral2} reviews related work on visual navigation and vision-language navigation. Section \uppercase\expandafter{\romannumeral3} introduces the problem formulation and details the architecture of VLFly. Section \uppercase\expandafter{\romannumeral4} presents experimental evaluations in both simulation and real-world environments. Section \uppercase\expandafter{\romannumeral5} concludes the paper and outlines future research directions.

\section{Related Work}
\subsection{Traditional Methods}
Traditional methods for visual navigation, including SLAM \cite{mur2015orb, matsuki2024gaussian} and SfM \cite{cui2023mcsfm, wang2024tc}, have become well-established through decades of research. Recently, many approaches have shifted toward using monocular RGB cameras \cite{mur2015orb, rodriguez2024nr}, which enable the reconstruction of environmental geometry and simultaneous ego-motion estimation. These approaches typically rely on handcrafted feature matching, incremental map building, and classical planners for obstacle avoidance and navigation. For instance, the system developed by Mur-Artal et al. \cite{mur2015orb} uses handcrafted ORB features for matching and tracking, and leverages a covisibility graph for local mapping. It is capable of real-time visual navigation in both indoor and outdoor environments. Despite their effectiveness in low-level navigation and mapping tasks, these approaches are fundamentally limited in their ability to interpret high-level semantic intent or incorporate natural language input. As such, they are not well-suited for VLN, where understanding and executing human instructions are essential for collaborative interaction.

\subsection{End-to-End Learning}
Current end-to-end learning methods utilize deep neural networks to directly map raw sensory inputs to output actions. These models are typically optimized via backpropagation using loss functions derived from IL \cite{ross2013learning, zhang2024npe, loquercio2021learning} or RL \cite{yan2021deep, xiao2023collaborative, yan2023collision, zhang2023partially, kaufmann2023champion, xiao2024learning, zhang2025learning}. For instance, Zhang et al. \cite{zhang2025learning} proposed a cross-modal contrastive learning-based framework for end-to-end monocular UAV navigation, which aims to extract environment-invariant features from RGB images and feed them into a downstream deep reinforcement learning (DRL) policy network for action prediction. Loquercio et al. \cite{loquercio2021learning} introduced an end-to-end vision-based high-speed navigation method that learns to map noisy sensory observations to collision-free trajectories by imitating an oracle teacher with access to full state information.

Despite the promising performance, end-to-end learning methods face several bottlenecks. These include low sample efficiency during early training, a high demand for extensive annotated data, and a lack of interpretability. Most notably, since such methods are predominantly trained in simulation, their generalization capability is severely limited due to the visual domain gap, simplified physics, and the absence of real-world uncertainties, often resulting in significant performance degradation when deployed in real-world environments.

\subsection{Modular Learning}
Modular learning methods \cite{wu2022image, gervet2023navigating, roth2024viplanner, chaplot2020object} aim to combine the strengths of both traditional and end-to-end approaches by replacing specific components within classical navigation pipelines with learnable counterparts. This paradigm allows models to incorporate high-level semantic priors while retaining the interpretability and engineering reliability of traditional architectures. For instance, Wu et al. \cite{wu2022image} proposed a hierarchical modular framework for image-goal navigation, which separates mapping, long-term goal prediction, and motion planning into distinct components. Learning is selectively applied to modules such as goal prediction, while traditional SLAM and planners are preserved, enabling more targeted learning and improved sim-to-real generalization. Similarly, Gervet et al. \cite{gervet2023navigating} introduced a modular vision-based navigation framework that maintains a classical pipeline while replacing the heuristic exploration module with a learned semantic exploration policy based on Mask R-CNN. This approach demonstrated superior transferability from simulation to real-world environments.

While modular learning frameworks benefit from decomposing the navigation task and extracting high-level visual representations from sensor inputs, they often rely heavily on real-world training data and are susceptible to cumulative errors across modules. Most importantly, existing methods are limited in their ability to leverage vision-language features and thus lack the capacity for human-level reasoning and instruction understanding.

\subsection{Vision-Language Navigation}
The rapid progress of LLMs and VLMs has further advanced the field of visual navigation by enabling agents to interpret natural language instructions and reason about visual scenes in a generalizable, task-agnostic manner. One key advantage of these foundation models is that they are pre-trained on large-scale datasets and can often be directly deployed to downstream navigation tasks without the need for learning from scratch, significantly reducing training costs and improving generalization.

% hirose2024lelan zheng2024towards
% an2024etpnav shah2023lm
VLN methods can also be broadly categorized into end-to-end learning \cite{zhang2024navid, xu2025flame, zhang2025mapnav} and modular learning \cite{yokoyama2024vlfm, wu2024voronav, du2025vl} approaches. End-to-end methods leverage pre-trained large models to extract aligned features from image and text inputs and directly map them to action outputs. By optimizing network parameters to approximate language-annotated reference trajectories, these approaches empower agents with semantic understanding and zero-shot navigation capabilities. For instance, Zhang et al. \cite{zhang2024navid} proposed an end-to-end VLN framework that integrates a pre-trained LLM to encode natural language instructions, which are combined with egocentric video observations to directly predict action commands. The policy is optimized using pre-recorded oracle trajectories as supervision. In contrast, modular learning approaches utilize pre-trained large models to extract semantic representations directly from raw sensor inputs, supporting zero-shot reasoning and spatial inference in VLN tasks. For instance, Yokoyama et al. \cite{yokoyama2024vlfm} introduced a modular VLN framework that combines traditional mapping and planning with pre-trained VLMs. It constructs a semantic value map by calculating the similarity between RGB observations and text prompts, and employs a frontier-based strategy to select the most promising waypoint to navigate next. 

However, most existing VLN methods, whether end-to-end or modular, are designed for ground-based robots and typically assume a discrete action space, which is ill-suited for aerial navigation. 
% However, UAVs differ substantially from ground robots in motion dynamics and observation spaces, making existing approaches ill-suited for aerial navigation. 
Although recent works \cite{liu2023aerialvln, lee2024citynav, gao2025openfly} have begun to explore the use of large models in VLN for UAVs, they often adopt ground-based settings and evaluate performance solely in simulation. Consequently, the applicability of these methods to real-world UAV VLN tasks with continuous control remains underexplored.

\begin{figure*}[!t]
   \centering
   \includegraphics[width=1.0\textwidth, height=0.35\textheight]{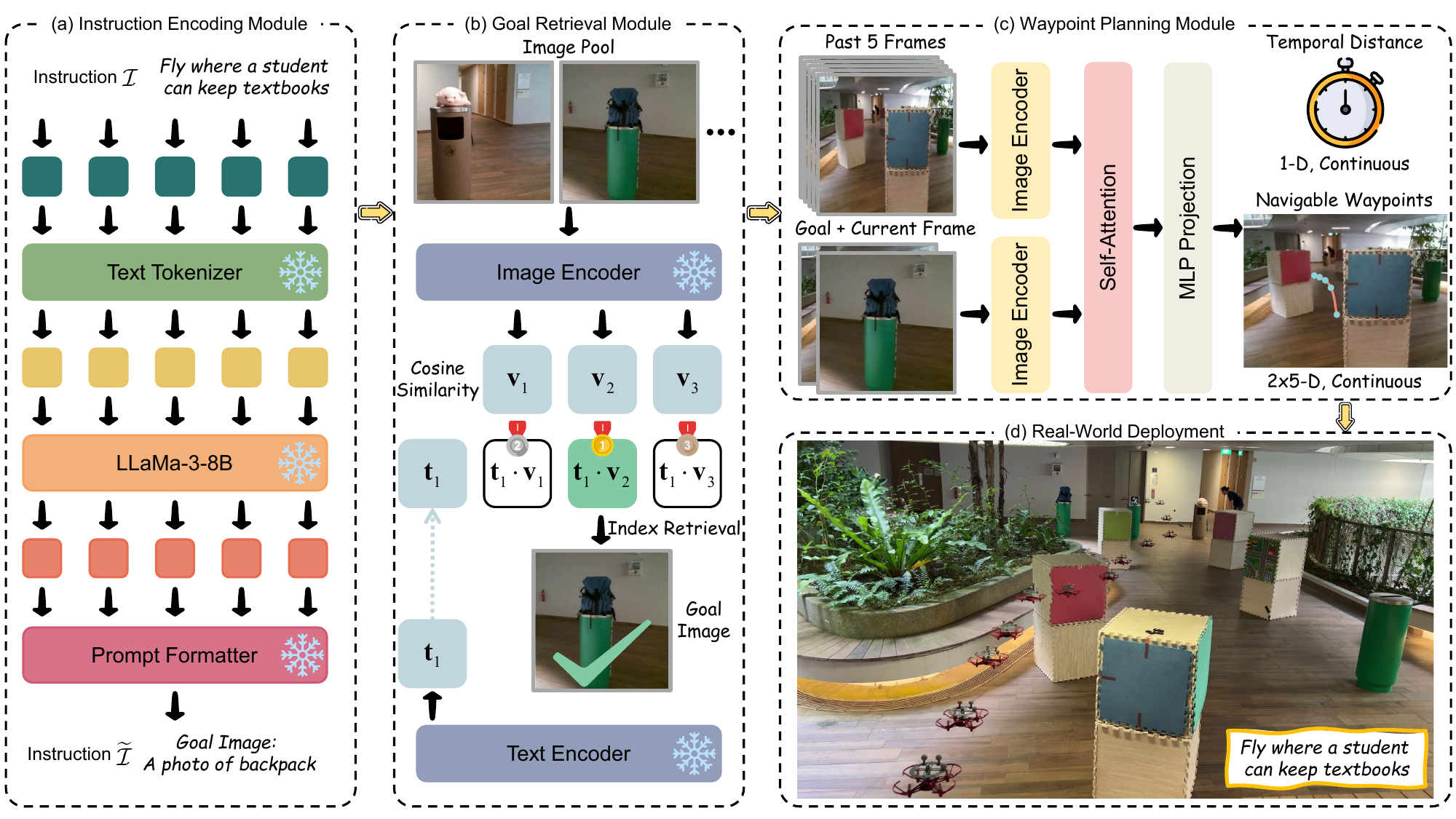} 
   \caption{The framework of VLFly. (a) an instruction encoder that reformulates natural language into structured prompts; (b) a goal retriever that selects the most relevant image via vision-language similarity; and (c) a waypoint planner that generates continuous trajectories from egocentric observations. These waypoints enable real-world UAV navigation without fine-tuning.}
   \label{fig:framework}
\end{figure*}

\section{Methodology}
\subsection{Problem Formulation}
The VLN task for UAVs in this work is formulated as a Partially Observable Markov Decision Process (POMDP), defined by the tuple $\langle \mathcal{S}, \mathcal{A}, \mathcal{O}, \mathcal{P} \rangle$. At each timestep $t$, the agent receives a high-level task instruction $\mathcal{I} = \{w_1, w_2, \dots, w_l\}$ consisting of $l$ natural language tokens, a goal image $\boldsymbol{I}^g$, and a visual observation $\mathcal{O}_t = \{\boldsymbol{I}_{t-k}, \dots, \boldsymbol{I}_t\}$, where each $\boldsymbol{I}_i \in \mathbb{R}^{H \times W \times 3}$ denotes an egocentric RGB frame captured by the UAV's onboard monocular camera. The agent selects an action $\boldsymbol{a}_t \in \mathcal{A}$, where $\mathcal{A} \subset \mathbb{R}^n$ defines a continuous control space, and each action $\boldsymbol{a}_t = (v_t, \omega_t)$ corresponds to linear and angular velocity commands. The action moves the agent to the next state according to the probabilistic state transition function $\mathcal{P}: \mathcal{S} \times \mathcal{A} \times \mathcal{S} \rightarrow[0,1]$, and the agent receives a new observation $\mathcal{O}_{t+1}$. Since the true environment state $\mathcal{S}$ is unobservable, the agent must rely solely on partial observations and semantic grounding to navigate toward the intended visual goal. Unlike prior methods \cite{liu2023aerialvln, lee2024citynav, gao2025openfly} that use discrete action sets composed of symbolic commands and fixed arguments, our approach adopts continuous velocity control, which is more aligned with real-world UAV behavior. 
% This formulation allows actions to be executed directly from visual input in real time, in a manner similar to how humans navigate with limited perceptual information.

\subsection{Overall Framework}
% We develop VLFly based on a series of task-agnostic foundation models, including LLaMA \cite{touvron2023llama}, CLIP \cite{radford2021learning}, and ViNT \cite{shah2023vint}. Rather than designing each model from scratch, we inherit the core architectures of these foundation models and introduce task-oriented framework-level design to adapt them to UAV-based VLN tasks. This design enables efficient knowledge transfer from general pre-training to our task-specific setting, making the generalization challenges more tractable.

As illustrated in Fig. \ref{fig:framework}, VLFly consists of three interconnected modules, namely an instruction encoding module, a goal retrieval module, and a waypoint planning module. Given a navigation instruction at timestep $t$, \textit{i.e.}, a natural language sentence describing the goal, the input instruction is first processed by LLaMA \cite{touvron2023llama} to produce a goal-oriented prompt, which explicitly encodes the intended visual concept in a standardized textual form. Subsequently, this prompt is matched against a pool of pre-collected candidate images leveraging CLIP \cite{radford2021learning}. The matching is performed by projecting both the prompt and candidate images into a shared embedding space aligned across modalities, where their similarity is measured to identify the goal image most consistent with the instruction. Once the goal image is determined, it is combined with the current egocentric observations and passed into the planning module, which produces a set of feasible waypoints guiding the UAV toward the goal. These waypoints are then translated into continuous velocity commands for real-time execution. Note that although VLFly comprises functionally distinct modules, the framework operates in an end-to-end manner at inference time, which maps the visual observation to action directly. Moreover, our work highlights novel framework-level modeling as detailed in the following.

\subsection{Module Details}
% \textbf{Instruction encoding module.} It is responsible for converting high-level natural language instructions into structured text prompts tailored for downstream goal image retrieval, as shown in Fig. \ref{}. Formally, given an input instruction \(\mathcal{I} = \{w_1, w_2, \dots, w_l\}\) consisting of \(l\) tokens, we first apply a tokenizer \(T(\cdot)\) to convert it into a sequence of discrete token IDs \(\mathbf{x} = T(\mathcal{I}) = \{x_1, x_2, \dots, x_l\},\ x_i \in \mathbb{Z}\). Each token \(x_i\) is then embedded into a continuous representation via an embedding matrix \(E(\cdot)\), resulting in \(\mathbf{e}_i = E(x_i),\ \mathbf{e}_i \in \mathbb{R}^d\), where \(d\) is the embedding dimension. The resulting embedding sequence \(\{\mathbf{e}_1, \dots, \mathbf{e}_l\}\) is fed into a pretrained language model \(f_{\text{LLM}}\) (e.g., LLaMA \cite{touvron2023llama}) to produce contextualized hidden states \(\mathbf{H} = f_{\text{LLM}}(\mathbf{e}_1, \dots, \mathbf{e}_l),\ \mathbf{H} \in \mathbb{R}^{l \times d}\). Based on the final hidden representation, the model autoregressively generates a task-specific prompt \(\tilde{\mathcal{I}} = \{w'_1, w'_2, \dots, w'_{l'}\}\), where each \(w'_j\) is drawn from the output vocabulary according to the distribution \(w'_j \sim p(w'_j \mid w'_1, \dots, w'_{j-1}, \mathcal{I})\). The generated textual prompt \(\tilde{\mathcal{I}}\) is then used as a query input for the subsequent module to identify the most semantically relevant goal image.

\subsubsection{Instruction encoding module} 
\begin{figure}[t!]
    \centering
    \includegraphics[width=0.9\linewidth]{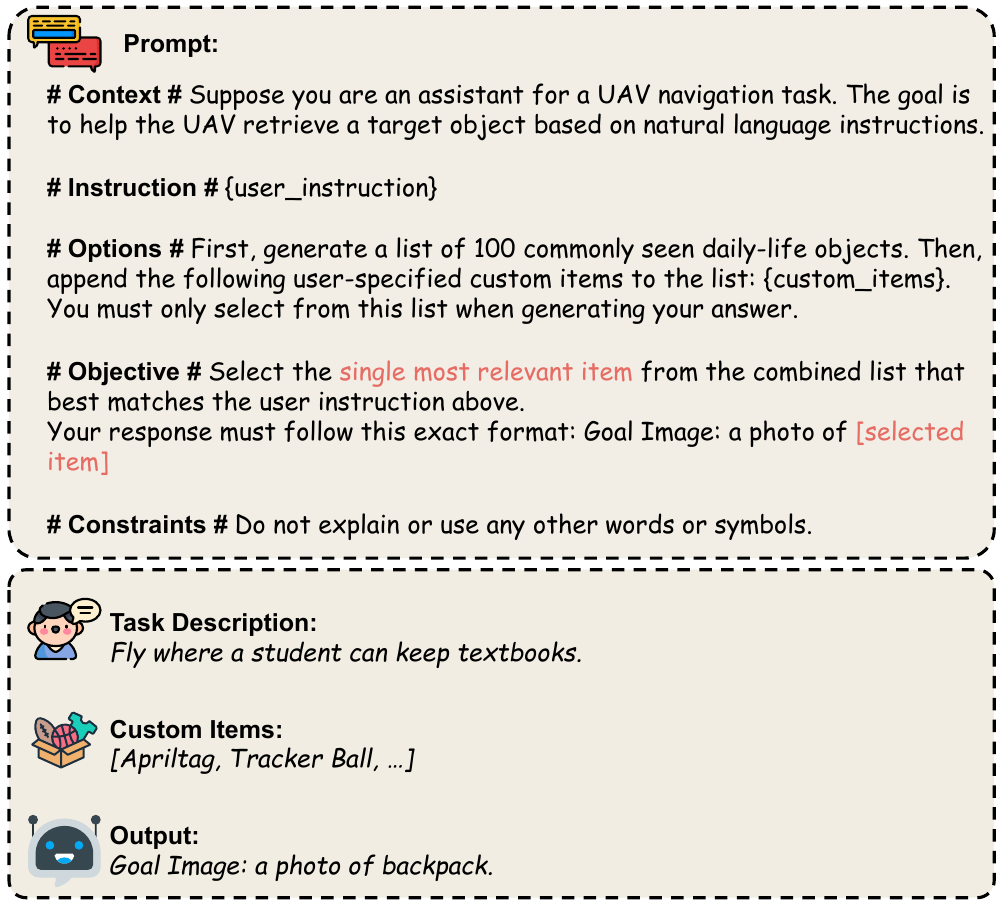}
    \caption{Prompting strategy used in the instruction encoding module. Given a natural language instruction and a list of custom items, the language model generates a standardized prompt in a specific format, enabling consistent goal retrieval.}
    \label{fig:reasoning}
\end{figure}

It aims to convert high-level natural language instructions into structured text prompts tailored for downstream goal image retrieval, as shown in Fig. \ref{fig:reasoning}. Formally, given an input instruction \(\mathcal{I} = \{w_1, w_2, \dots, w_l\}\) consisting of \(l\) tokens, we first apply a tokenizer \(T(\cdot)\) to convert it into a sequence of discrete token IDs: 
\begin{equation}
\mathbf{x} = T(\mathcal{I}) = \{x_1, x_2, \dots, x_l\}, \quad x_i \in \mathbb{Z}.
\end{equation}

Each token \(x_i\) is then embedded into a continuous representation via an embedding matrix \(E(\cdot)\), resulting in \(\mathbf{e}_i = E(x_i),\ \mathbf{e}_i \in \mathbb{R}^d\), where \(d\) is the embedding dimension. The resulting embedding sequence \(\{\mathbf{e}_1, \dots, \mathbf{e}_l\}\) is fed into a pre-trained language model \(f_{\text{LLM}}\) (e.g., LLaMA~\cite{touvron2023llama}) to produce contextualized hidden states:
\begin{equation}
\mathbf{H} = f_{\text{LLM}}(\mathbf{e}_1, \dots, \mathbf{e}_l), \quad \mathbf{H} \in \mathbb{R}^{l \times d}.
\end{equation}

Based on the final hidden representation, the model autoregressively generates a task-specific prompt \(\tilde{\mathcal{I}} = \{w'_1, w'_2, \dots, w'_{l'}\}\), where each token is sampled according to:
\begin{equation}
w'_j \sim p(w'_j \mid w'_1, \dots, w'_{j-1}, \mathcal{I}).
\end{equation}

The generated textual prompt \(\tilde{\mathcal{I}}\) is then used as a query input for the subsequent module to identify the most semantically relevant goal image.

% \textbf{Goal retrieval module.} It aims to identify the most semantically aligned goal image from a predefined candidate pool based on the prompt generated in the previous stage. Specifically, we leverage CLIP \cite{radford2021learning} to encode both the prompt and the images into a shared embedding space. Given a prompt \(\tilde{\mathcal{I}}\), the text encoder \(f_{\text{T}}(\cdot)\) maps it to a normalized text embedding \(\mathbf{t} \in \mathbb{R}^d\). Similarly, each image \(\boldsymbol{I}_j\) in the candidate pool \(\mathcal{G} = \{\boldsymbol{I}_1, \boldsymbol{I}_2, \dots, \boldsymbol{I}_N\}\) is encoded by the image encoder \(f_{\text{V}}(\cdot)\) into a normalized visual embedding \(\mathbf{v}_j \in \mathbb{R}^d\). We then compute similarity scores between the prompt and each image as scaled dot products \(s_j = \tau^{-1} \cdot \langle \mathbf{t}, \mathbf{v}_j \rangle\), where \(\tau^{-1}\) is a learned logit scaling factor. To obtain a probability distribution over the candidate images, we apply a softmax over the similarity scores: \(p_j = \frac{\exp(s_j)}{\sum_{k=1}^{N} \exp(s_k)}\). Finally, the goal image \(\boldsymbol{I}^g\) with the highest similarity score is selected and subsequently leveraged together with egocentric observations to generate waypoints.

\subsubsection{Goal retrieval module} It is responsible for identifying the most semantically aligned goal image from a predefined candidate pool based on the prompt generated in the previous stage. Specifically, we leverage CLIP~\cite{radford2021learning} to encode both the prompt and the images into a shared embedding space. Given a prompt \(\tilde{\mathcal{I}}\), the text encoder \(f_{\text{T}}(\cdot)\) maps it to a normalized text embedding \(\mathbf{t} \in \mathbb{R}^d\). Similarly, each image \(\boldsymbol{I}_j\) in the candidate pool \(\mathcal{G} = \{\boldsymbol{I}_1, \boldsymbol{I}_2, \dots, \boldsymbol{I}_N\}\) is encoded by the image encoder \(f_{\text{V}}(\cdot)\) into a normalized visual embedding \(\mathbf{v}_j \in \mathbb{R}^d\). We then compute similarity scores between the prompt and each image as scaled dot products:
\begin{equation}
    s_j = \tau^{-1} \cdot \langle \mathbf{t}, \mathbf{v}_j \rangle,
\end{equation}
where \(\tau^{-1}\) is a learned logit scaling factor. To obtain a probability distribution over the candidate images, we apply a softmax over the similarity scores:
\begin{equation}
p_j = \frac{\exp(s_j)}{\sum_{k=1}^{N} \exp(s_k)}.
\end{equation}

Finally, the goal image \(\boldsymbol{I}^g\) with the highest similarity score is selected and subsequently leveraged together with egocentric observations to generate waypoints.

\subsubsection{Waypoint planning module} It takes a sequence of egocentric observations from the current and past \(P\) timesteps as inputs, along with the goal image, and outputs (i) the estimated number of steps needed to reach the goal (the temporal distance \(\hat{d} \in \mathbb{R}_+\)), and (ii) a trajectory of \(H\) future relative waypoints \(\hat{\boldsymbol{a}} = \{\hat{\boldsymbol{a}}_1, \dots, \hat{\boldsymbol{a}}_H\}\), where each \(\hat{\boldsymbol{a}}_i \in \mathbb{R}^n\) is the expected egocentric displacement of the UAV at future step \(i\). 
To process temporal visual information, each observation image \(\boldsymbol{I}_{t-k} \in \mathbb{R}^{H \times W \times 3}\) (\(k = 0, \dots, P\)) is independently encoded using an image encoder \(\psi(\cdot)\), producing a flattened feature vector \(\psi(\boldsymbol{I}_{t-k}) \in \mathbb{R}^d\). These \(P+1\) feature vectors are stacked to represent the observation context. In parallel, to capture the relative semantics between the current observation and the goal image, we adopt a goal fusion encoder \(\phi(\boldsymbol{I}_t, \boldsymbol{I}^g)\), where the current observation \(\boldsymbol{I}_t\) and goal image \(\boldsymbol{I}^g\) are concatenated and fed into the encoder. The output is a flattened goal token \(\phi(\boldsymbol{I}_t, \boldsymbol{I}^g) \in \mathbb{R}^d\), which serves to guide the navigation process by encoding goal-directed difference features rather than absolute goal content. The \(P+1\) observation embeddings and the goal token are concatenated with positional encodings before being passed into a decoder-only Transformer backbone \(f(\cdot)\). The output sequence is then processed by an MLP head to predict the final outputs.

Importantly, unlike the original ViNT \cite{shah2023vint} which was trained on heterogeneous ground robot datasets, we apply this architecture in the UAV-based VLN setting, where egocentric observations exhibit higher viewpoint variability and spatial diversity than those in ground-based environments. These domain characteristics introduce unique visual navigation challenges, suggesting the robustness of the integrated planning module within our framework.

\subsubsection{Action execution.} 
Given the predicted continuous trajectory from the waypoint planning module, where each waypoint is normalized to the range \([-1, 1]\), we implement a simple PID controller to convert each waypoint into executable action commands. Specifically, the two components of each predicted waypoint \(\hat{\mathbf{a}}_i = (x_i, y_i)\) are first linearly scaled by:
\begin{equation}
    (x_i, y_i) \leftarrow (x_i, y_i) \cdot \left(\frac{v_{\max}}{f_c}\right),
\end{equation}
where \(v_{\max}\) denotes the maximum allowable linear velocity and \(f_c\) is the control frequency. The unnormalized displacement is then passed to a PID controller, which maps the 2D offset to continuous linear and angular velocities. These high-level commands are directly executed by the UAV in real time.

\subsection{Implementation Details}
We evaluate VLFly in both simulated and real-world environments without any additional training. All modules are deployed using their default pre-trained parameters. In simulation, the framework is tested in Unity, running on a workstation equipped with an NVIDIA GeForce RTX 4090 GPU. The system operates at a control frequency of approximately 15\,Hz
 in the simulated environment. For real-world experiments, VLFly is executed on a laptop with an NVIDIA GeForce RTX 4070 GPU. In this setting, the system receives RGB images captured by the onboard UAV camera along with text instructions and directly outputs action commands, which are transmitted remotely for execution by the UAV. The overall control frequency in the real-world setup is approximately 7--10\,Hz, depending on communication latency and onboard sensing conditions.

\section{Simulations and Experiments}

\begin{figure}[t!]
    \centering
    \begin{subcaptionbox}{Furniture scenario\label{fig:tra_furniture}}[0.7\linewidth]
        {\includegraphics[width=\linewidth]{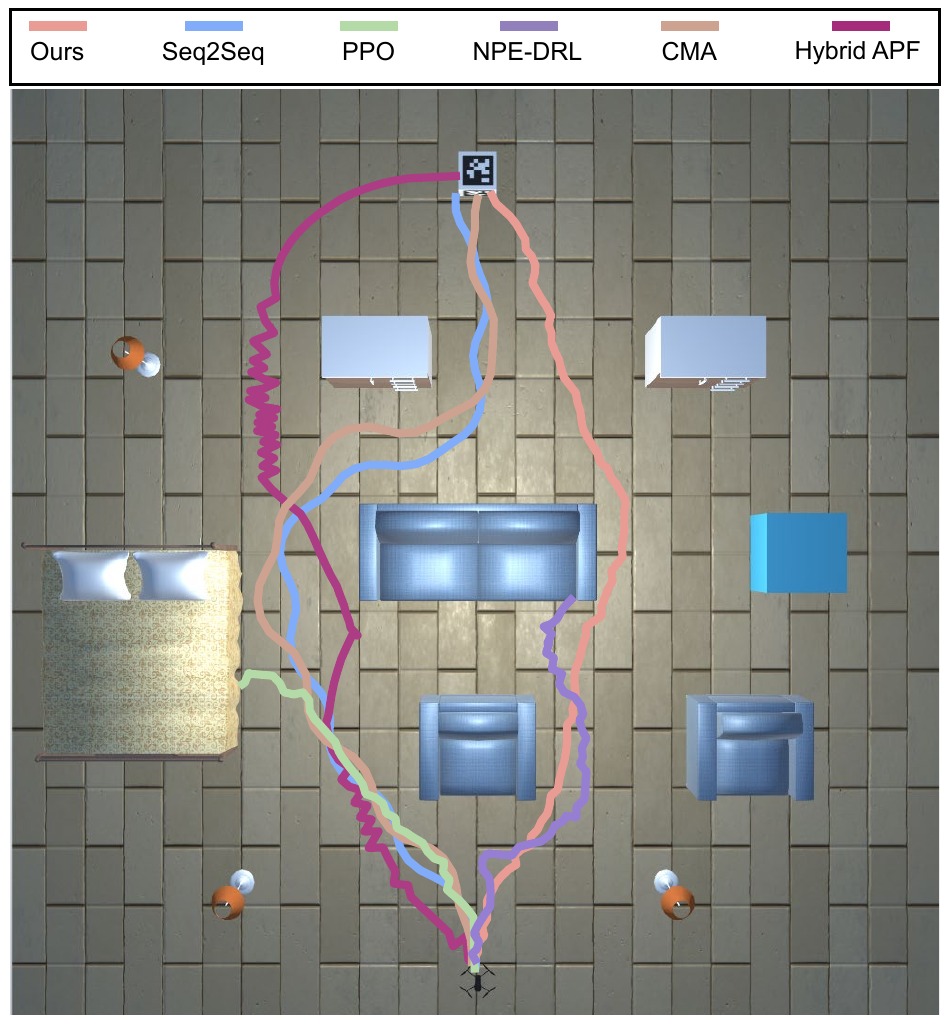}}
    \end{subcaptionbox}
    
    \vspace{1em} % 垂直间距

    \begin{subcaptionbox}{Barrier scenario\label{fig:tra_barrier}}[0.7\linewidth]
        {\includegraphics[width=\linewidth]{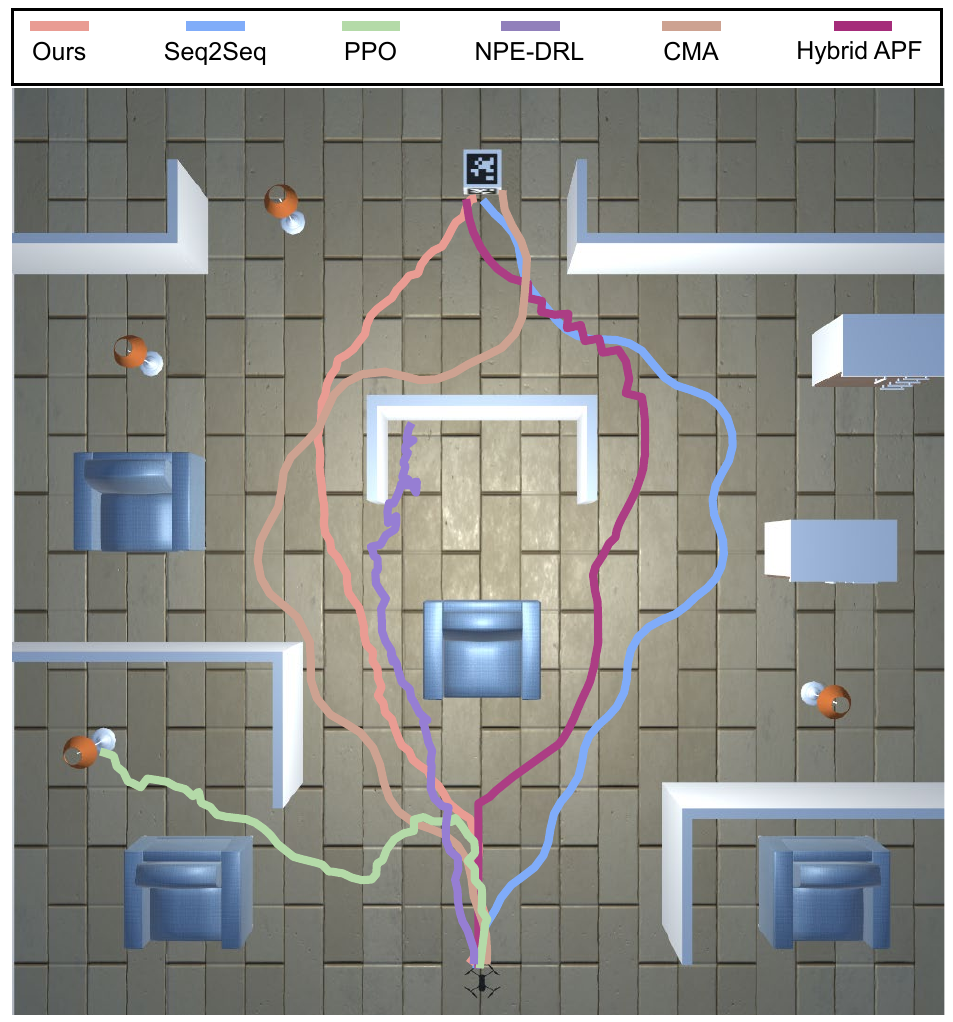}}
    \end{subcaptionbox}
    
    \caption{Comparison of navigation trajectories across different baselines in simulation. VLFly consistently produces the most robust and smooth trajectories in both scenarios, outperforming all baselines in goal-directedness and obstacle avoidance. Note that the box environment is excluded, as it is primarily used for training certain baselines.}
    \label{fig:tra}
\end{figure}

\subsection{Metrics and Baselines}

We adopt commonly used metrics in VLN \cite{anderson2018vision} to assess navigation performance, including Success Rate (SR), Oracle Success Rate (OS), Success weighted by Path Length (SPL), and Navigation Error (NE). SR denotes the percentage of episodes in which the agent reaches the goal, while OS measures whether any point along the trajectory falls within the success threshold $\delta$, which is set to $0.5\,\mathrm{m}$. SPL reflects task success and path efficiency, and is computed as $\text{SPL} = \frac{1}{N} \sum_{i=1}^{N} \frac{S_i \cdot l_i}{\max(p_i, l_i)}$, where $S_i \in \{0,1\}$ indicates success, $l_i$ is the shortest path length, and $p_i$ is the actual path length. NE is defined as the average Euclidean distance (in meters) from the agent's final position to the goal. An episode is considered successful if the UAV reaches the target in simulation or arrives within $0.8\,\mathrm{m}$ of the goal in real-world experiments.

To thoroughly evaluate the performance of VLFly, we compare it against several representative baselines. Seq2Seq \cite{anderson2018vision} is a basic sequence-to-sequence baseline that employs a recurrent policy to predict action primitives from expert demonstrations based on RGB-D inputs. In our implementation, we only use monocular images as input for a fair comparison. PPO \cite{schulman2017proximal} is a widely used RL baseline that optimizes action policies through trial-and-error interaction with the environment. The policy takes RGB images as input and directly outputs continuous action commands. NPE-DRL \cite{zhang2024npe} is a hybrid learning framework that combines non-expert IL with RL to better capture the pixel-to-action mapping. It leverages sub-optimal demonstrations to guide exploration and accelerate policy learning. CMA \cite{anderson2018vision} is a cross-modal attention-based VLN model that jointly processes visual observations and natural language instructions. To adapt it to our task, we preserve its original recurrent decoding structure while projecting the predicted navigable actions into fixed linear and angular velocities. Hybrid-APF \cite{pan2021improved} is a classical path planning baseline based on the artificial potential field (APF) method, where the agent's movement is guided by the attractive force of the goal and the repulsive force of obstacles.

\subsection{Comparison in Simulated Environments}
We construct our simulation environments in Unity to comprehensively evaluate the performance and generalization of VLFly. To this end, we design three distinct environments of increasing complexity: a simple box environment, a moderately complex furniture environment, and a challenging cluttered environment with irregular barriers, as shown in Fig. \ref{fig:tra}. This progressive setup allows us to test the robustness and scalability of VLFly under varying degrees of perceptual and planning difficulty. Note that the box environment is used solely for training certain baselines and is excluded from trajectory-based evaluations, which are conducted primarily in the furniture and cluttered environments.

For a fair comparison, all baselines are evaluated solely on their navigation performance in simulation, without involving any language grounding components. In this setting, the UAV is tasked with reaching a specified goal position from the starting point without any collision using only egocentric visual input. The instruction-driven, open-vocabulary goal understanding and navigation capabilities are instead evaluated in real-world experiments.

Among all the baselines, PPO and NPE-DRL are trained in the simple box environment and directly evaluated in the furniture and barrier settings. Seq2Seq and CMA adopt the offline training paradigm proposed in \cite{anderson2018vision}, where pre-trained models are directly applied during testing. In contrast, Hybrid-APF and VLFly require no training and are deployed directly across all environments. Each baseline is evaluated over 200 episodes in each environment. The quantitative results are summarized in Table. \ref{tab:baseline}, and trajectories of each baseline are shown in Fig. \ref{fig:tra}.

\begin{figure*}[htbp]
    \centering
    \begin{subcaptionbox}{Trajectory visualization under different instructions\label{fig:indoor_tra}}[0.8\textwidth]
        {\includegraphics[width=\linewidth]{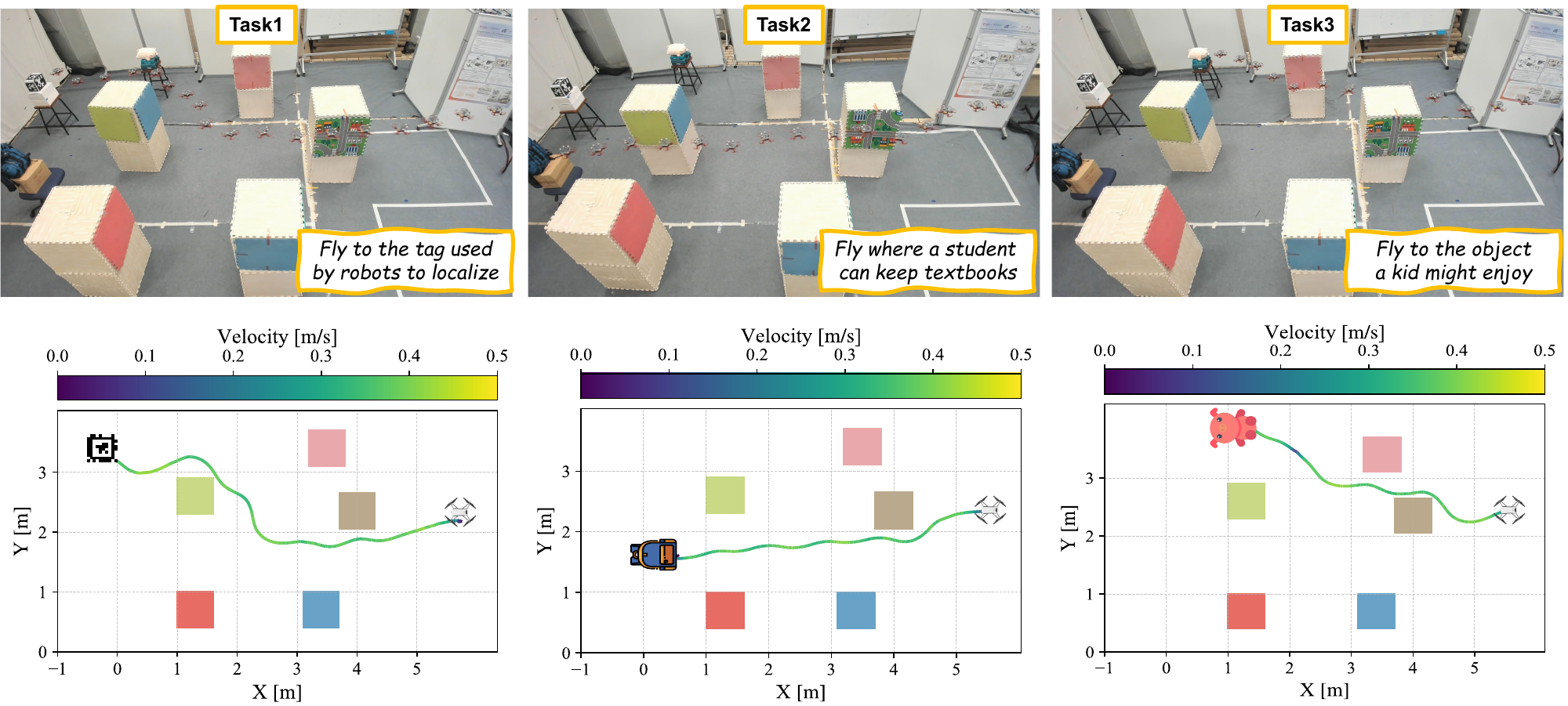}}
    \end{subcaptionbox}

    \vspace{1em}

    \begin{subcaptionbox}{Egocentric waypoint visualization showing evasive maneuvers and goal-oriented motion\label{fig:indoor_waypoint}}[0.8\textwidth]
        {\includegraphics[width=\linewidth]{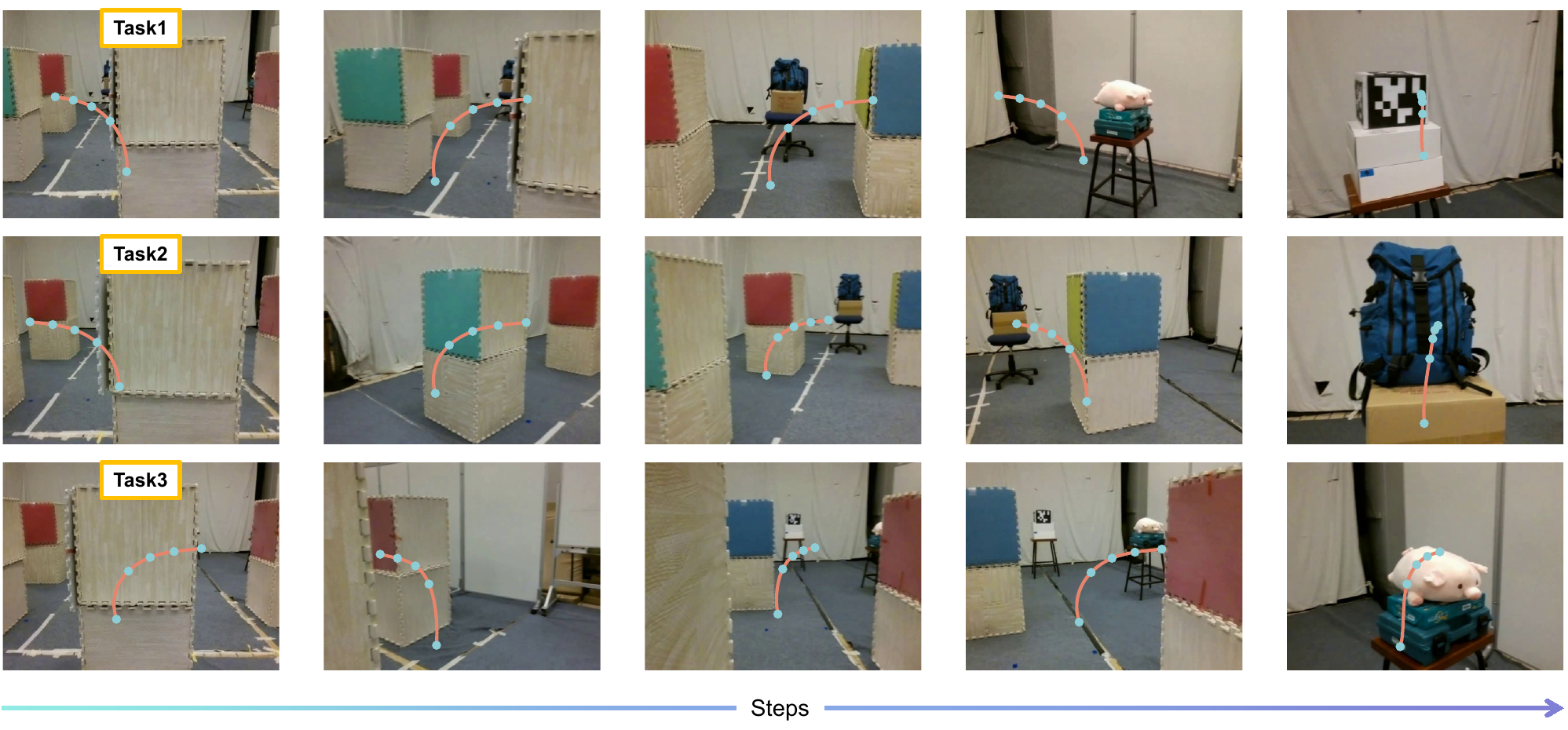}}
    \end{subcaptionbox}
    
    \caption{Real-world indoor flight results of VLFly under different natural language instructions. The UAV navigates to the correct target based solely on egocentric observations, demonstrating its ability to interpret abstract language and avoid obstacles through effective waypoint planning.}
    \label{fig:indoor}
\end{figure*}

% \begin{table}[!t]
% \caption{Performance of Various Baselines in the Furniture Environment}
% \label{table_furniture}
% \centering

% \renewcommand{\arraystretch}{1.0}
% \setlength{\tabcolsep}{6pt}

% \begin{tabular}{
%     c
%     S[table-figures-integer=1,table-figures-decimal=2,table-figures-uncertainty=0]
%     S[table-figures-integer=1,table-figures-decimal=2,table-figures-uncertainty=0]
%     S[table-figures-integer=2,table-figures-decimal=2,table-figures-uncertainty=2]
%     S[table-figures-integer=2,table-figures-decimal=2,table-figures-uncertainty=2]
% }
% \toprule
% \multirow{2}{*}{Method} & \multicolumn{4}{c}{Furniture} \\
% \cmidrule(lr){2-5}
%  & {NE~$\downarrow$} & {OS~$\uparrow$} & {SR~$\uparrow$} & {SPL~$\uparrow$} \\
% \midrule
% Ours      & \num{10.95}~$\pm$~\num{0.73} & 0.0 & 0.0 & 0.0 \\
% Seq2Sqe \cite{anderson2018vision}                   & \num{3.12(1.73)} & 52.0 & 40.5 & 38.7 \\
% PPO \cite{schulman2017proximal}        & \num{9.80(1.73)} & 0.0 & 0.0 & 0.0 \\
% NPE-DRL \cite{zhang2024npe}     & \num{3.20(0.94)} & 49.2 & 39.8 & 36.5 \\
% CMA \cite{anderson2018vision} & 3.50 & 0.0 & 0.0 & 0.0 \\
% Hybrid APF \cite{pan2021improved}   & 3.35 & 48.0 & 38.6 & 36.1 \\
% \bottomrule
% \end{tabular}
% \end{table}

VLFly achieves the best performance across all metrics, except in the box environment where some baselines were trained. However, VLFly still performs competitively and significantly outperforms all baselines in unseen environments. This is attributed to the zero-shot reasoning capability of foundation large models and the grounding of navigation intent in high-level semantic features. Seq2Seq and CMA, though trained on high-fidelity datasets, suffer from limited adaptability. Their pretraining paradigm is confined to narrow environmental distributions, making them prone to overfitting to known scene layouts and visual representations. As a result, they exhibit notably reduced performance in unseen environments. Similarly, PPO and NPE-DRL perform well in the training environment but are particularly vulnerable to distribution shifts. Their policies are tightly coupled with the training environment, leading to poor generalization in unseen domains due to overfitting and limited sample efficiency. Hybrid-APF shows decent performance in structured environments such as the box and furniture scenarios, where geometric regularity benefits potential field calculations. However, it easily falls into local minima in more complex and cluttered scenarios, leading to significant performance degradation.

\subsection{Real-World Experiments}
To evaluate the real-world performance of VLFly, we deploy it in indoor and outdoor environments without any additional fine-tuning. In the indoor setting, we construct a test scenario with multiple cube-shaped obstacles of varying textures and colors. Distinct goal objects, including a backpack, a toy, and an AprilTag, are individually placed at different locations. The UAV used for testing is a Tello Edu, equipped with a monocular RGB camera operating at 720p and 30 FPS, with a field of view (FOV) of $82.6^\circ$. The task requires the UAV to navigate toward the object specified in the instruction while avoiding collisions. The outdoor setting introduces a more complex environment with greater spatial openness, dynamic lighting variations, and increased background clutter such as dense vegetation. Compared to the indoor setup, it imposes additional challenges in visual perception and control stability. Importantly, all real-world experiments are conducted without reliance on external localization systems such as OptiTrack. The entire navigation process is driven solely by egocentric RGB observations to validate VLFly's open-vocabulary goal understanding and generalized navigation capabilities.

To further validate VLFly's open-vocabulary goal understanding and generalized navigation capabilities, we conduct extensive physical flight experiments in various challenging real-world environments. Two types of instructions are designed to comprehensively evaluate system performance: direct and indirect. Direct instructions explicitly describe the target object (e.g., ``\textit{fly to the blue backpack}''), whereas indirect instructions contain high-level abstract semantics (e.g., ``\textit{fly where a student can keep textbooks}''), requiring additional reasoning to infer the intended goal.

\begin{figure*}[htbp]
    \centering
    \begin{subcaptionbox}{Trajectory visualization under different instructions\label{fig:outdoor_tra}}[0.8\textwidth]
        {\includegraphics[width=\linewidth]{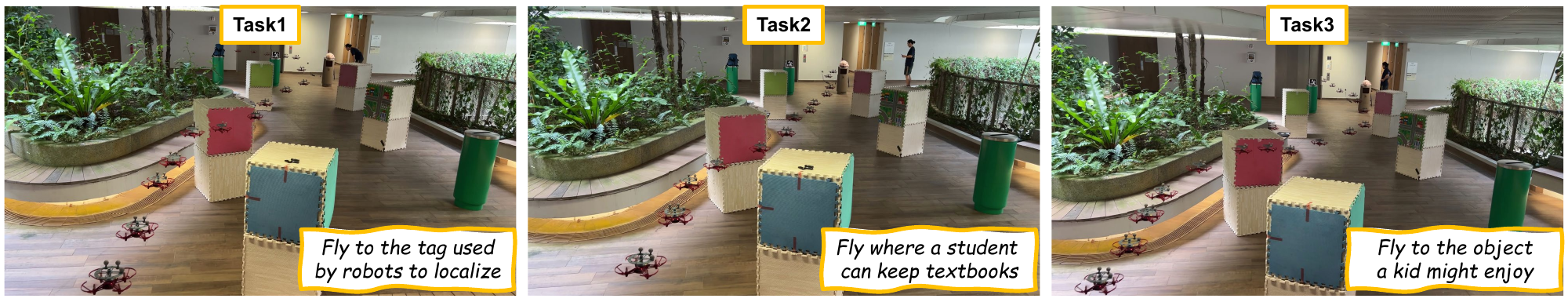}}
    \end{subcaptionbox}

    \vspace{1em}

    \begin{subcaptionbox}{Egocentric waypoint visualization showing evasive maneuvers and goal-oriented motion\label{fig:outdoor_waypoint}}[0.8\textwidth]
        {\includegraphics[width=\linewidth]{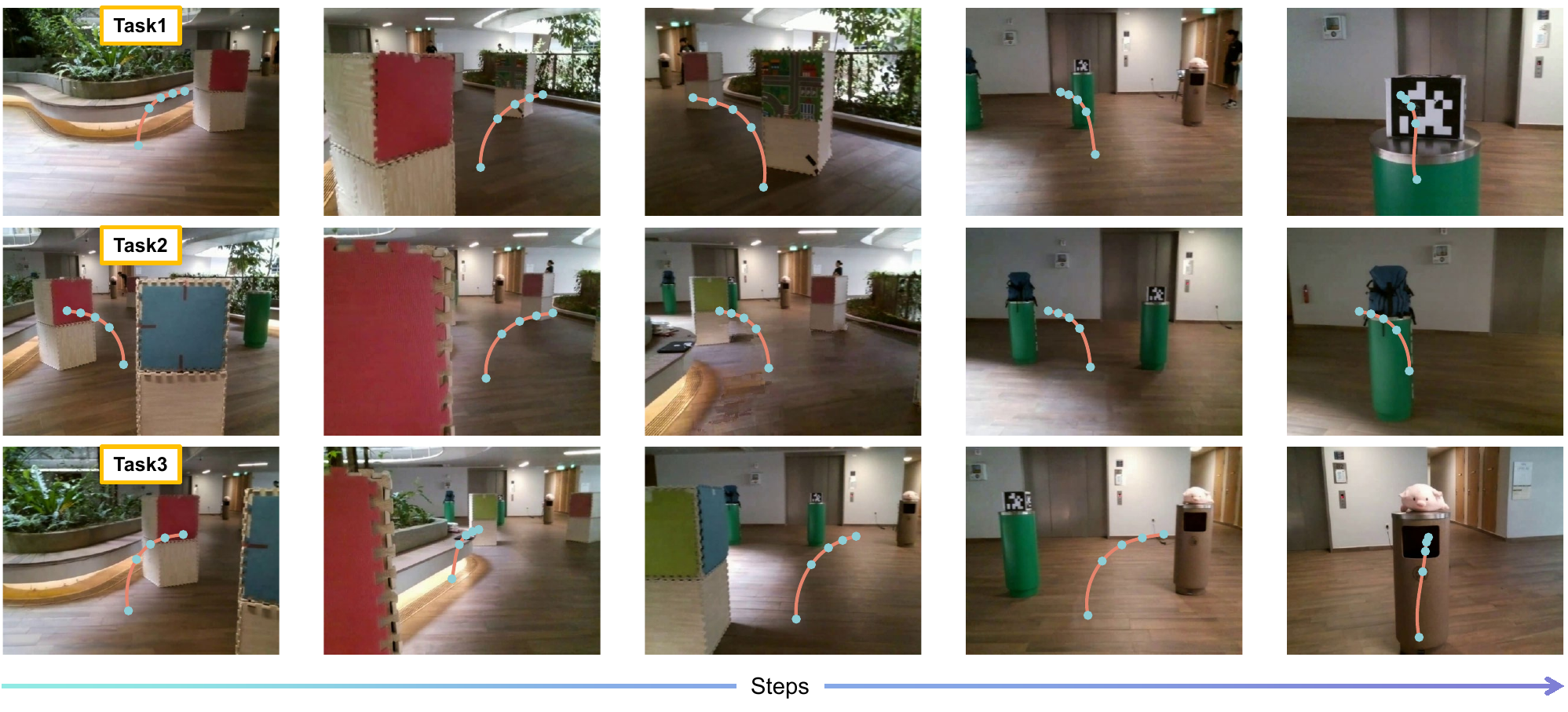}}
    \end{subcaptionbox}
    
    \caption{Real-world outdoor flight results of VLFly under different natural language instructions. Despite increased lighting variation and background clutter, the UAV successfully navigates to the correct targets based solely on egocentric observations, demonstrating robust instruction grounding and obstacle-aware trajectory generation.}
    \label{fig:outdoor}
\end{figure*}

Both instruction types are evaluated with 20 trials each in both indoor and outdoor settings. VLFly achieves a success rate of \(83\%\) for direct instructions and \(70\%\) for indirect ones, using only monocular RGB image sequences as input. These results demonstrate VLFly's ability to overcome the sim-to-real gap caused by lighting variations, texture shifts, and discrepancies in UAV dynamics. The performance of VLFLy is attributed to the robust semantic grounding enabled by the instruction encoding and goal retrieval modules. The reduced success rate for indirect instructions stems from inherent semantic ambiguity, where different object categories (e.g., bookshelf and backpack) may plausibly fulfill the same instruction, leading to incorrect goal grounding. To offer intuitive insight, we visualize the flight trajectories and the predicted waypoints from the UAV's first-person view, as shown in Figs. \ref{fig:indoor} and \ref{fig:outdoor}. These results demonstrate that VLFly can effectively guide the UAV around obstacles toward the intended goal.

\begin{table*}[!t]
\caption{Performance of Various Baselines in Simulation Environments}
\label{tab:baseline}
\centering
\renewcommand{\arraystretch}{1.1}
\begin{tabular*}{\textwidth}{@{\extracolsep{\fill}}lcccccccccccc}
\toprule
\multirow{2}{*}{Method} & 
\multicolumn{4}{c}{Box Scenario (Easy)} & 
\multicolumn{4}{c}{Furniture Scenario (Medium)} & 
\multicolumn{4}{c}{Barrier Scenario (Hard)} \\
\cmidrule(lr){2-5} \cmidrule(lr){6-9} \cmidrule(lr){10-13}
& NE$\downarrow$ & OS$\uparrow$ & SR$\uparrow$ & SPL$\uparrow$
& NE$\downarrow$ & OS$\uparrow$ & SR$\uparrow$ & SPL$\uparrow$
& NE$\downarrow$ & OS$\uparrow$ & SR$\uparrow$ & SPL$\uparrow$ \\
\midrule
VLFly & 1.57 & 88.8 & 86.4 & 0.82
             & \textbf{1.73} & \textbf{84.6} & \textbf{82.5} & \textbf{0.79}
             & \textbf{2.06} & \textbf{81.7} & \textbf{77.3} & \textbf{0.75} \\
Seq2Seq~\cite{anderson2018vision} & 4.65 & 37.4 & 35.1 & 0.32
                                  & 5.89 & 24.5 & 21.2 & 0.19
                                  & 6.17 & 12.3 &  8.9 & 0.06 \\
PPO~\cite{schulman2017proximal}   & 1.35 & 92.3 & 90.4 & 0.88
                                  & 6.90 & 17.6 & 11.8 & 0.08
                                  & 7.36 & 11.9 & 0.0 & 0.00 \\
NPE-DRL~\cite{zhang2024npe}       & \textbf{1.02} & \textbf{94.2} & \textbf{92.2} & \textbf{0.90} 
                                  & 6.08 & 21.5 & 17.6 & 0.12
                                  & 7.59 & 9.7 & 0.0 & 0.00 \\
CMA~\cite{anderson2018vision}     & 4.28 & 39.1 & 38.9 & 0.36
                                  & 5.50 & 27.9 & 25.8 & 0.23
                                  & 6.94 & 18.7 & 10.2 & 0.08 \\
Hybrid APF~\cite{pan2021improved} & 2.42 & 75.0 & 73.0 & 0.70
                                  & 2.56 & 73.5 & 72.0 & 0.69
                                  & 5.23 & 31.1 & 28.7 & 0.25 \\
\bottomrule
\end{tabular*}
\end{table*}

\begin{table*}[htbp]
\caption{Ablation Study of VLFly in Simulation Environments}
\label{tab:ablation}
\centering
\renewcommand{\arraystretch}{1.1}
\begin{tabular*}{\textwidth}{@{\extracolsep{\fill}}lcccccccccccc}
\toprule
\multirow{2}{*}{Method} & 
\multicolumn{4}{c}{Box Scenario (Easy)} & 
\multicolumn{4}{c}{Furniture Scenario (Medium)} & 
\multicolumn{4}{c}{Barrier Scenario (Hard)} \\
\cmidrule(lr){2-5} \cmidrule(lr){6-9} \cmidrule(lr){10-13}
& NE$\downarrow$ & OS$\uparrow$ & SR$\uparrow$ & SPL$\uparrow$
& NE$\downarrow$ & OS$\uparrow$ & SR$\uparrow$ & SPL$\uparrow$
& NE$\downarrow$ & OS$\uparrow$ & SR$\uparrow$ & SPL$\uparrow$ \\
\midrule
VLFly & 1.57 & 88.8 & 86.4 & 0.82 
             & \textbf{1.73} & \textbf{84.6} & \textbf{82.5} & \textbf{0.79}
             & \textbf{2.06} & \textbf{81.7} & \textbf{77.3} & \textbf{0.75} \\
w/o Prompting & 3.46 & 52.8 & 49.5 & 0.47
                                  & 4.33 & 38.6 & 36.4 & 0.34
                                  & 4.82 & 35.1 & 32.3 & 0.30 \\
w/ Unified VLM   & 3.82 & 47.4 & 43.8 & 0.40
                                  & 4.57 & 38.6 & 36.3 & 0.33
                                  & 4.95 & 33.9 & 30.8 & 0.28 \\
w/ RL Policy       & \textbf{1.14} & \textbf{93.7} & \textbf{91.6} & \textbf{0.89}
                                  & 6.52 & 19.3 & 13.1 & 0.10 
                                  & 7.68 & 8.9 & 0.0 & 0.00 \\
\bottomrule
\end{tabular*}
\end{table*}

\subsection{Ablation Studies}
To verify the effectiveness of each module in VLFly, we conduct ablation studies by removing or replacing specific components and measuring their impact on overall performance. Three ablated variants are designed:
\begin{itemize}
    \item w/o Prompting: The instruction encoding module is removed. Raw language instructions are directly used as input for goal retrieval without prompt generation.
    \item w/ Unified VLM: The instruction encoding and goal retrieval modules are jointly replaced with a unified VLM (e.g., BLIP \cite{li2022blip}). This model performs end-to-end matching between instruction and candidate goal images.
    \item w/ RL Policy: The waypoint planning module is replaced with a RL-based policy network. Instead of predicting waypoints, it directly outputs action commands based on the current image and retrieved goal.
\end{itemize}

All ablation experiments are conducted in the simulation environments with indirect language instructions. Quantitative results are presented in Table \ref{tab:ablation}. Across all metrics, each variant shows noticeable performance degradation. Specifically, w/o Prompting shows a substantial drop under indirect instruections. This is because removing the instruction encoding module significantly weakens the system's reasoning capacity. The goal retrieval module alone is only effective when the input instruction conforms to fixed textual patterns (e.g., ``\textit{a photo of...}''), and struggles to interpret abstract semantics. W/ Unified VLM similarly fails to handle indirect instructions effectively. Although unified VLMs such as BLIP  \cite{li2022blip} excel at image-conditioned reasoning and multi-turn visual question answering, they lack precise control over prompt structure and typically yield ambiguous responses instead of interpretable similarity scores. Additionally, such models require iterating over the entire image pool with separate forward passes, resulting in prohibitive inference latency. These limitations render them unsuitable as drop-in replacements for the modular instruction encoding and goal retrieval pipeline in VLFly. W/ RL Policy maintains accurate goal retrieval but performs poorly in unseen environments due to the limited generalization ability of the policy network. Compared to the structured waypoint planning module, the RL policy suffers from insufficient training exposure and is highly sensitive to distributional shifts. As a result, it tends to overfit training trajectories and fails to generate effective evasive maneuvers in complex environments.

Overall, each module in VLFly is indispensable. The instruction encoding module transforms natural language into structured prompts, the goal retrieval module grounds these prompts to semantic goals, and the waypoint planning module generates structured, interpretable control commands. Only through this cascaded design can VLFly achieve robust and generalizable VLN performance.

\section{Conclusion}
In this paper, we propose VLFly, a novel VLN framework designed for UAVs. VLFly achieves outstanding navigation using only egocentric RGB observations, without relying on external localization or active ranging sensors. Distinct from existing VLN paradigms, VLFly outputs continuous control commands rather than selecting from a discrete action set, enabling better UAV maneuverability. Specifically, the framework consists of three modules: an instruction encoder that converts natural language into structured text prompts, a goal retriever that selects the best-matching image based on vision-language similarity, and a waypoint planner that generates interpretable trajectories from egocentric observations. Extensive simulation experiments validate that VLFly consistently outperforms all baselines across various environments. Moreover, real-world flight tests in both indoor and outdoor scenarios demonstrate that VLFly retains open-vocabulary goal understanding and generalization under direct and indirect instructions, confirming its practical viability for real-world VLN task execution.

In the future work, VLFly can be improved from the following aspects. First, the waypoint planning module relies on structural consistency across sequential RGB inputs, which constrains UAV motion to a two-dimensional plane without altitude control. Introducing training datasets with a broader action space can support full three-dimensional maneuvering. Furthermore, the goal retrieval module depends on a predefined image pool, limiting adaptability in open-world environments. To overcome this, future research will explore the potential of VLMs to dynamically identify goal candidates from real-time observations by integrating them with object detection or segmentation techniques.

% References

\bibliographystyle{Bibliography/IEEEtranTIE}
\bibliography{Bibliography/yuhang}\ %IEEEabrv instead of IEEEfull

\vspace{-1cm}

\end{document}